# A Novel ANN Structure for Image Recognition

Shilpa Mayannavar, Uday Wali, and V M Aparanji

Shilpa Mayannavar is a research scholar at C-Quad Research Desur IT Park, Belagavi. She is a full time PhD student at Visvesvaraya Technological University, Belagavi – 590018, Karnataka India. Phone: +91 8431478910; e-mail: mayannavar.shilpa@gmail.com

Uday Wali is a CEO of C-Quad Computers, Desur IT Park, Belagavi-590014. He is an adjunct professor at Dept. of Electronics and Communication and Engineering, SGBIT, Belagavi-590010, Karnataka India.

V M Aparanji is Associate Professor at Dept. of Electronics and Communication Engineering, SIT, Tumakuru-572103

**ABSTRACT**
The paper presents Multi-layer Auto Resonance Networks (ARN), a new neural model, for image recognition. Neurons in ARN, called Nodes, latch on to an incoming pattern and resonate when the input is within its 'coverage.' Resonance allows the neuron to be noise tolerant and tunable. Coverage of nodes gives them an ability to approximate the incoming pattern. Its latching characteristics allow it to respond to episodic events without disturbing the existing trained network. These networks are capable of addressing problems in varied fields but have not been sufficiently explored. Implementation of an image classification and identification system using two-layer ARN is discussed in this paper. Recognition accuracy of 94% has been achieved for MNIST dataset with only two layers of neurons and just 50 samples per numeral, making it useful in computing at the edge of cloud infrastructure.

## 1. INTRODUCTION

Recently, Deep Learning (DL) systems have been very successful in solving complex problems like image recognition, robotic motion control and natural language processing. Some of the popular DL systems include the Convolutional Neural Networks (CNNs) [1] [2] for image recognition, Long Short-Term Memory (LSTM) [3] for robotic control and time series prediction, Generative Adversarial Networks (GANs) [4] for image synthesis etc.

Deep Learning systems are computationally intensive. Graphics Processing Units (GPUs) have shouldered most of this burden [5] till now. Several software tools and platforms like Caffe [6], Theano [7] and TensorFlow [8] have been developed around GPUs. However, several new platform specific hardware designs have been available to implement DL applications. Cambricon [9], IBM TrueNorth [10] and Google TPU [11] are some examples of such hardware accelerators for DL systems. Multi-layer Artificial Neural Networks (ANNs) are at the core of all DL systems. Complexity of DL systems arises from the large number of inputs and interdependency of system parameters, generally called as the curse of dimensionality [12]. ANNs implementing such DL systems, therefore, tend to be very sparse: Significant part of the computation may not contribute to final outcome of the DL system. Most of the neural networks do not clearly demarcate the neuronal paths or their dependencies in any particular recognition. So, pruning the networks becomes difficult. It should be possible to increase the overall performance of ANN by increasing the density of the network.

Neural networks in biological systems consist of various kinds of neurons, each performing a specific function. Role of specific neurons depends on various factors like their structure, position in the network, connectivity, dendrite density, length of axon, neurotransmitters and inhibitors used, chemical receptors and gateways, type of input and output, timing response and a plethora of other factors. It is therefore important to explore various neural architectures to move towards realization of anything closer to Artificial General Intelligence (AGI). Further evolution in artificial neural networks has to see several new types of neural architectures evolving, generalizing as well as specializing in the functionality of a neuronal type. The necessity of refining existing deep learning neural models is being reported in recent literature, e.g., capsule networks [13] and spiking networks [14].

Auto Resonance Networks (ARN) proposed by Aparanji et al. [15] for use in robotic path planning using graphs, has some interesting properties in this direction. Each node in ARN is analogous to a biological neuron capable of specific recognition function. ARN has two distinct learning mechanisms: (a) The nodes learn by tuning their resonance characteristics in response to variations in input and (b) layers of ARN learn by identifying spatial and temporal associations between neuronal outputs in lower layers or input data. Output of every node in ARN is limited to an adjustable but finite upper bound. This alleviates the stability problems typically seen in Hebbian-like associative learning systems. As the output is constrained to an upper bound, long strands of patterns may be identified by multi layer ARN. Both learning modes impart an ability to dynamically morph the solution space in response to variations in input. Further, activation paths established by the network corresponding to specific input-output relations can be traced and explained. Therefore, it is easy to find how and why ARN performed a particular task. This makes it is easy to prioritize nodes and paths that significantly contribute to overall functioning of the network. In turn, this allows pruning of the network to increase computational efficiency of the recognition system. Temporal behavior of ARN and its application has also been discussed in [15]. However, no implementation details were presented. Use of ARN in other applications has not been well documented. In this paper, we have reported implementation details of an image identification/ classification system based on hierarchical ARN. One of the standard benchmarks used in image classification is the hand written character dataset in MNIST library [16][17], which has been used in this work. Implementation details, elaborations and results are presented in this paper.

Section 2 of this paper presents an overview of ARN and its adaptation for image identification/classification. Section 3 discusses implementation details. Results are presented in section 4. Discussion on other applications that may benefit from hierarchical ARN are presented in section 5.

## 2. AUTO RESONANCE NETWORK

Basic neuronal structure in Auto Resonance Network (ARN) is different than that used in other ANN. Therefore, an overview of the model is made here. A typical biological neuron has several dendrites that receive inputs from sensor neurons or intra neurons in its vicinity. The strength of the dendrite connection presents a scaled version of its input to the neuron. These scaled inputs are added to generate an internal state of excitation. When the level of excitation exceeds a threshold, the neuron fires output. The output is related to the level of excitation but the physical transport mechanism puts an upper limit on the output. Therefore the output is a saturating non linear function of the sum of scaled inputs. This also implies that there is a certain noise tolerance built into this neuronal excitation: Variations in inputs can still result in the same level of excitation. Connection strength of individual inputs to the neuron varies with repeated use, which forms the basis of constrained Hebbian learning. Level of excitation is indicated by the time through which the excited state is held.

### 2.1 Constrained Hebbian Learning

Hebbian rule states that the strength of a dendritic connection improves as a particular input is applied more frequently or reduces if not used. The rule is somewhat partial as it does not clearly state how the output of a neuron is limited to a saturation level or the length of time the neuron stays in excited state. Most of the current interpretations of Hebbian rule use it without any constraint on the strength of the connection. This leads to the classical weakness of Hebbian learning, i.e., continued use of a connection will scale the input so much that the node becomes unstable. In a biological system, this never happens as the physical limits on transport will limit the strength of connection and hence the Hebbian rule must be subjected to an upper limit. ARN implements such a constrained Hebbian learning mechanism that limits the output of a node to a threshold. Characteristics of ARN nodes are discussed in following sections.

### 2.2 The Resonator

Consider a simple resonator like
$$y = x * (k - x) \tag{1}$$
It has a resonance (peak) value of $y_m = k^2/4$ occurring at $x_m = k/2$. Both $y_m$ and $x_m$ are independent of x. Therefore, this peak value remains unaffected by translation, scaling or other monotonic transformations of x like sigmoid or hyperbolic tangent. It is possible to shift the point of resonance by translating the input to required value by translation or scaling.

For translation, $X \Leftarrow x - p$, peak occurs at
$$x_m = (k/2) + p \tag{2}$$
For scaled input, $X \Leftarrow xp$, peak occurs at
$$x_m = k/(2p) \tag{3}$$
In either case, peak value is bound by
$$y_m = k^2/4 \tag{4}$$

This is an important feature as the output is bound irrespective of how x is scaled by the strength of connection. Though k may be used for gain control, using fixed value per layer or network yields stable systems. We have used $k = 1$ giving a peak value of ¼. Thus output may be multiplied by 4 to yield a peak value of 1. It is also possible to use $k = 2$ for peak value of 1.

Several envelop function to transform input are discussed in [18]. One of these interesting input transformations is the sigmoid:
$$X = \frac{1}{1+e^{-\rho(x-x_m)}} \tag{5}$$
where ρ, called the resonance control parameter, is a positive real number and $x_m$ is the resonant input. For now, we will assume $k = 1$. Note that $(1 - X)$ has form similar to equation (5) and is given by
$$(1 - X) = \frac{1}{1+e^{\rho(x-x_m)}} \tag{6}$$
Substituting (5) and (6) in (1), we get
$$y = X(1-X) = \frac{1}{(1+e^{-\rho(x-x_m)})} \frac{1}{(1+e^{\rho(x-x_m)})} \tag{7}$$

The output of a resonator expressed by equation (7) attains a peak value of $y_m = 1/4$ at $X_m = 1/2$ as expected. Incidentally, equation (7) is also the derivative of equation (5) which has a well known bell shaped curve.

Figure 1 shows some node outputs for $k = 1$. Figure 1(a) and Figure 1(b) show shifting resonance by input translation and input scaling given by equation (2) and (3) respectively. Figure 1(c) shows the output curves for sigmoid transform for different values of $x_m$. Resonance of each shifted sigmoid in Figure 1(c) can be altered by using resonance control parameter ρ, as shown in Figure 1(d). These equations allow constrained Hebbian learning to be incorporated into ANNs without the associated instability problems. It must be emphasized that our usage of the term constrained Hebbian learning is only a holistic description of the ARN's learning algorithm. Actual learning algorithm may be a variation of Hebbian learning, suitable for specific use case.

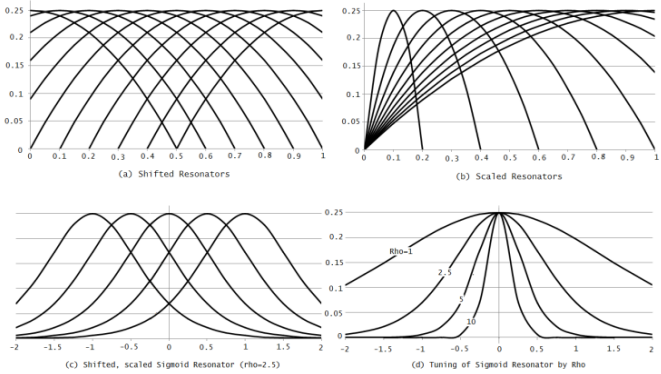

**Figure 1.** Resonating curves of ARN nodes (k = 1)

### 2.3 Selecting the Resonance Parameters

The location of the peak and the maximum value for equation (1) are controlled by the only parameter k. In case the input is transformed, the effect of k on the resonance should be studied. For example, equation (5) transforms real input x to X. Corresponding ARN in equation (7) has a peak of ¼ occurring at $X_m = k/2 = 1/2$ or equivalently $x = x_m$. In order to scale the peak to 1, we may use k = 2. However, $X_m$ shifts to 1 and X does not cross 1 for any x. Therefore, the resonator can only implement left side of the resonance function. This suggests that choice of k must ensure that the reverse transformation $x = f^{-1}(X)$ will yield x on either side of $X_m = k/2$. Conversely, k should be selected such that x is distributed on either side of $x_m$. It is not necessary that the transformation is symmetric about $X_m$ but must have values spread on either side. For example, Figure 2 shows the effect of transformation given in equation (5). As k shifts, $X_m$ and $x_m$ also shift. There may be other parameters like ρ in equation (7) that are specific to the chosen transformation which also need to be checked for validity of the input range.

### 2.4 Threshold, Coverage and Label of a Node

A node is said to be triggered if the output is above a pre-selected threshold T. Several input points in the neighborhood of the peak will cause the node to be triggered. The set of input values to which the node is triggered is called the coverage of the node. As mentioned earlier, node learns by adjusting the coverage in response to the input. No two nodes will have same coverage. Coverage of two or more nodes may overlap only partially, which can cause several nodes to be triggered for some inputs. If only one node fires, network has identified the input and hence can classify it. If the input is covered by more than one node, one of the nodes may adjust its coverage to produce stronger output. If no node produces an output above threshold T, then a new node is created. Clearly, the number of nodes in ARN varies depending on the type of inputs, training sequence, resonance control parameter (ρ) and threshold (T).

The output of a node is a constrained real value, which may be input to the nodes in higher layers. Nodes may also be labeled with a class which reflects the type of primary input. Many nodes may be tagged with the same label. Therefore, a label may indicate an arbitrary collection of nodes. This allows input space to be continuous or disconnected, convex or concave, linear or non-linear, etc. This feature of ARN layers gives them an ability to learn from any input set.

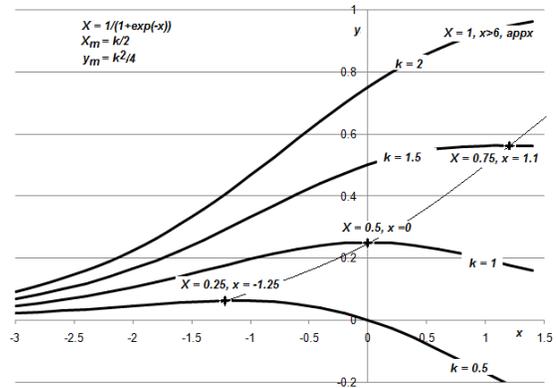

**Figure 2.** Dependence of resonance on k for equation (5)

### 2.5 The Aggregator

A node in ARN can have several inputs, each with its own resonator and a distinct coverage. Assuming that suitable transformation has been applied, output of the node with N inputs is given by, say

$$y = \frac{4}{Nk^2} \sum_{i=1}^{N} X_i (k - X_i) \qquad (8)$$

with a peak value of 1. The term $4/Nk^2$ is used to normalize the output to 1. If this output is above a threshold, the node gets triggered and therefore identifies the input.

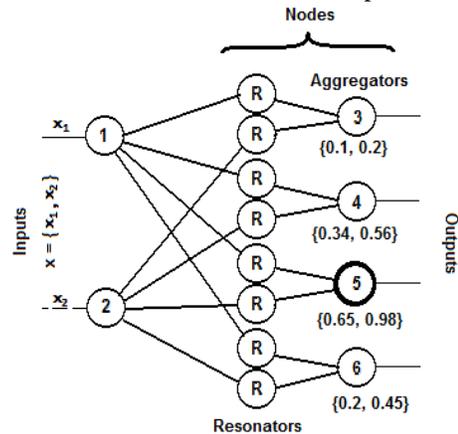

**Figure 3.** A layer of ARN structure

### 2.6 Constructing an ARN

In a biological system, neurons develop during early childhood but the connections are made on learning. It is possible to start an ARN with randomized weights but a computationally efficient ARN may be implemented as a dynamic network which grows with variations in applied input: The connections are simultaneously the cause and effect of learning. As an illustration, consider Figure 3. On initialization, only inputs 1 and 2 are present, representing the axonal receptacles from the primary sensor or outputs of other layers. The ARN layer is empty on initialization. Nodes get appended to the layer as the inputs are applied. For illustration, consider the first input {0.1, 0.2} applied to input nodes 1 and 2 shown in Figure 3. There will be no output from the network as there are no output nodes, implying no recognition. This causes node 3 to be created as first node in the layer. The second input {0.34, 0.56} is outside the coverage of node 3 and hence the network does not produce any output again. Another node, node-4 is created to resonate at this input and appended to the layer. The process continues on arrival of unmatched input. An input of {0.55, 0.86} is within the coverage of node 5 and hence, it fires. The winner node is highlighted with thicker lines. The two and three dimensional views of coverage of each output node are shown in Figure 4.

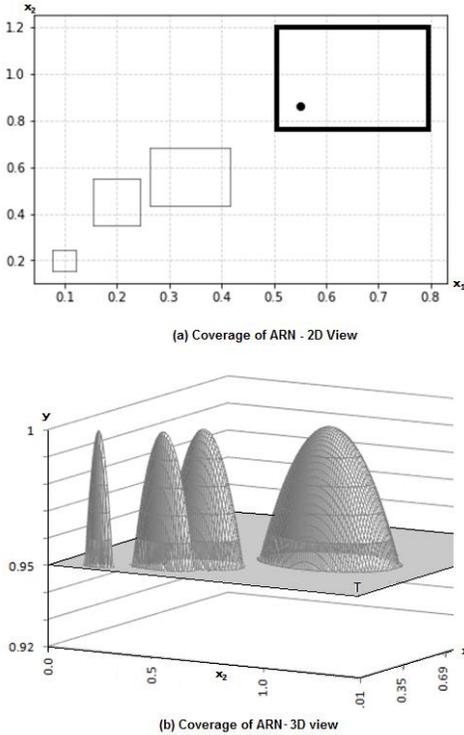

**Figure 4.** Coverage of sample ARN (a) 2D view (b) 3D view

The rectangles in Figure 4(a) represent the coverage of each output node. There are four rectangles corresponding to four output nodes of ARN shown in Figure 3. Each node is tuned to one unique input and has its own coverage. For example, the rectangle in thicker line representing the node 5 has the coverage area bound by the range $0.5 < x_1 < 0.8$ and $0.79 < x_2 < 1.2$. The test input {0.55, 0.86} shown as black dot (see Figure 4(a)) is within the coverage area of node 5 and therefore it fires.

### 2.6 Input Range, Coverage and Tuning

When a new node is created, default coverage is assigned to the node. Under certain transforms, coverage depends on point of resonance, as in Figure 4. In order to better control the learning in ARN, it is necessary to relate the statistical parameters of the input to the resonance control parameter and selection threshold of ARN nodes. It is possible to assume some reasonable values when a node is created but it would be more interesting if there is a basis for the estimate. Typical procedure to calculate initial coverage for a node is given below. The method may be suitably modified for other transformations.

One reasonably good threshold point is half power point, i.e.,

$$\text{T} = \text{half power point} = \sqrt{\frac{0.25^2}{2}} = 0.176 \quad (9)$$

Assuming $k = 1$ and the output of a node to be equal to threshold (T), we can equate the equation (1) to T and write

$$X(1 - X) = T = 0.176 \quad (10)$$

Let us use scaled input model of Figure 1(b) as an example, i.e., $X = xp$. By solving for x we get two points $x_c$ indicating bounds of coverage points:

$$x_c = \frac{1 \pm \sqrt{(1-4T)}}{2p} \quad (11)$$

Both $x_c$ points are symmetrically placed around the resonant point, which is given by equation (3) to be $x_m = 1/(2p)$. This is also evident from Figure 1(b).

Similar evaluation can be done for sigmoid transform. From equation (7), assuming $x_m = 0$, we can write

$$0.176 = \frac{1}{(1+e^{-\rho x})} \frac{1}{(1+e^{\rho x})} \quad (12)$$

Simplifying, and substituting $(e^{\rho x} + e^{-\rho x})/2 = \cosh(\rho x)$, we get

$$x_c = \frac{\pm \cosh^{-1}(1.8409)}{\rho} = \frac{\pm 1.2198}{\rho} \quad (13)$$

Therefore, as the value of $\rho$ increases, coverage reduces and vice-versa. The graph of $\rho$ Vs. $x_c$ is shown in Figure 5.

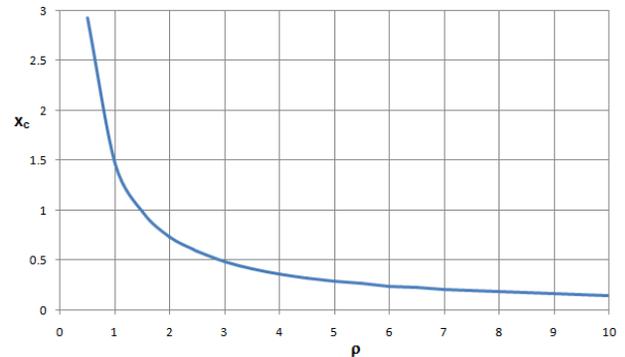

**Figure 5.** $x_c$ as a function of resonance control parameter, $\rho$ from equation (13)

Input range between the two $x_c$ is the coverage. Coverage for each of the inputs i.e., $x_1, x_2$ needs to be computed. The coverage and resonance control parameter depend on the transform used. Reduced coverage indicates focused learning. It is also helpful in resolving ambiguities in learning: In case of multiple winners, one of the nodes can relax control parameters to increase the coverage and enhance the output.

### 2.7 Data Variance and Node Coverage ($\rho$)

As coverage is a data dependant parameter, we may try to relate it to the statistical distribution of values arriving at a node. Assuming that the incoming data has a Gaussian distribution, we can derive a relation between the statistical variance of the data and coverage of ARN node. The Gaussian distribution is given by equation (14).

$$y = \frac{1}{\sqrt{2\pi\sigma^2}} e^{\frac{-(x-x_m)^2}{2\sigma^2}} \quad (14)$$

Where, $x_m$ indicates resonant input and $\sigma^2$ is variance. At $x = x_m$, we get the maximum value of y as

$$y_{peak} = \frac{1}{\sqrt{2\pi\sigma^2}} \quad (15)$$

Value of Gaussian distribution function at half power point can be written as

$$y_c = \sqrt{\frac{(y_{peak})^2}{2}} = \frac{1}{2\sqrt{\pi\sigma^2}} \quad (16)$$

Equating equations (14) and (16) and substituting $x_m = 0$, which is similar to the condition for equation (13), we get,

$$x_c = \pm 0.8325\sigma \quad (17)$$

In equation (17), $x_c$ represents the range of inputs (coverage of a node) and $\sigma$ represents standard deviation. From equations (13) and (17) we can get the relation between $\rho$ and $\sigma$;

$$\rho = \frac{\pm 1.4652}{\sigma} \quad (18)$$

When the mean and standard deviation are known, we can compute the corresponding values of $x_c$ and $\rho$. This in turn yields the coverage of the node. Graph in Figure 6 describes the coverage of node in range {0, 1} at $x_m = 0.5$. For other values of $x_m$ the coverage in terms of standard deviation can be expressed as

$$x_c = x_m \pm \alpha\sigma \quad (19)$$

where $\alpha$ indicates a scaling factor, similar to the constant in equation (17).

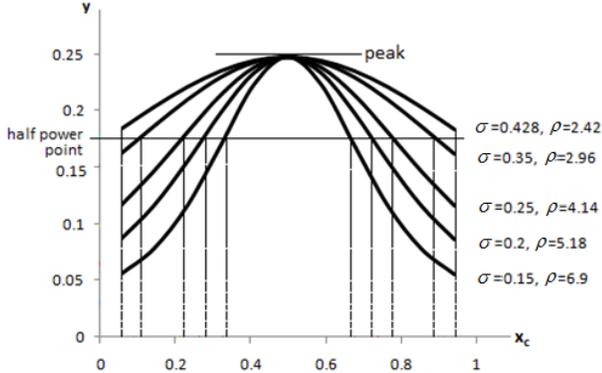

**Figure 6.** Tuning ARN Nodes

## 3. AUTO RESONANCE NETWORK BASED IMAGE RECOGNITION

Image recognition and classification has become a routine task using Convolutional Neural Networks (CNNs). However, there is still room for improvement. Capsule networks suggested by Hinton et al. [13] incorporate a spatial relation among the features recognized by CNNs. Sensitivity of CNN to quantization has been explored in [19]. Computational complexity of CNNs is very large, requiring support of large servers with special purpose accelerators. High performance implementations suitable for mobile devices are also being explored [20]. This situation hints at the necessity to explore alternate architectures to address the image recognition and classification problems. In this section, we will demonstrate the capability of ARN for Image Recognition.

For a quick proof of concept implementation, the network is trained using public MNIST database of handwritten digits [16]. It has a collection of 60,000 training samples and 10,000 test samples. As ARNs are noise tolerant, small number of training samples is sufficient. Set of 50 randomly chosen samples of each digit, making a total of 500 samples were used for training. The performance of ARN was evaluated using 150 test samples, representing about 30% of training set size. An accuracy of about 93% was observed. Details of this implementation are presented below. Larger datasets were used after the algorithm was validated.

### 3.1 Architecture

Essentially there are two layers of ARN marked as L1 and L2. L1 receives parts of image as input and converts them in a feature index. These recognized indices are temporarily stored in a spatially ordered list. This list is applied to L2, which will recognize the digit. Nodes in L2 are labeled with the class of image identified by the layer. During supervised training, the nodes are marked with specific class labels. For MNIST database, the labels are the values of digits, viz., zero to nine.

Tiling may be used to break the image into parts, to be input to L1, which recognizes them by identifying a matching part index. The sequence of recognized features represents a spatial relation between the features of input image. If the order of tiles is changed, the spatial relation between the features is altered and hence the output from the classifier may differ. For example the tiles may be numbered 1 to 16 from top left to bottom right. Presenting them to L2 in reverse order rotates the image by 180 degrees. Similarly, spatial reordered lists can be interpreted as mirror, shift, rotate etc. These operators are not significant for digit recognition but may improve recognition in several other use cases. Effectively, reordering of the feature list can serve as an internal synthesizer. Layer L2 can be trained with these altered lists to improve recognition accuracy.

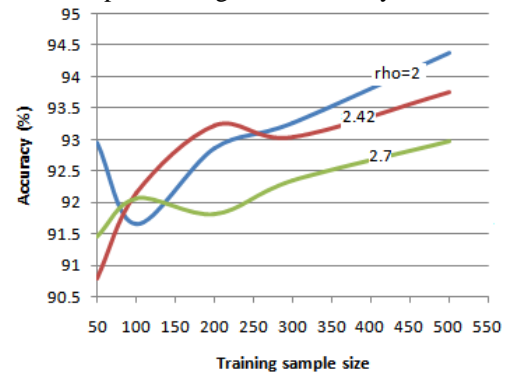

**Figure 7.** Training sample size Vs Recognition Accuracy

We have trained the network for different training sample size viz., 50, 100, 200, 300, 500. As the number of training samples increases, the accuracy of recognition also increases. Tunability of ARN nodes makes it possible to achieve an accuracy up to 93% with very few training samples as 50x10. Both $\rho$ and $T$ have effect on learning, as shown in Figure 7 and Figure 8: for the given dataset, $\rho = 2.42$ and $T = 0.9$ gave better overall results. These observations are discussed later in section 3.4.

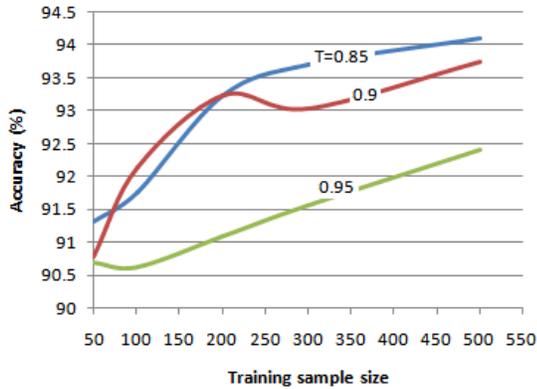

**Figure 8.** Effect of threshold T on learning

### 3.2 Effect of Perturbation

Perturbation refers to introduction of small changes to the system. In a network with large number of nodes, appending new nodes that are related but different from existing nodes is equivalent to a learning step. Perturbed nodes may require validation during later stages of learning but it provides an opportunity to respond well to an unknown input. Recognition accuracy of ARN can be improved by adding perturbed nodes to the network, albeit at the cost of increased computation. Perturbation can occur in input, output or network parameters. Perturbation may be random but limited to small variations such that the estimated response is not wrong but within a limited range of acceptability. In some cases, it may even be possible to use locally correct analytical equations to synthesize perturbed nodes. Adding a perturbation layer to ARN, will also reduce the number of images required to train the network. Some of the samples after rotating the image by small angles are shown in Figure 9. Aparanji et al., in a paper on generation of path for robotic motion using ARN [21], have reported that results may be improved by perturbing output and system parameters.

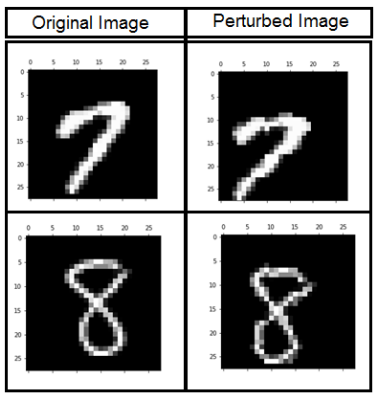

**Figure 9.** Example of perturbed images

### 3.3 Effect of Input Masking

The number of resonators can be reduced if we can use sparse connections instead of complete image tile. For MNIST images, it was reasonable to avoid resonators which were tuned to black pixels and keep only those with pixel values greater than some predefined limit. This could result in reduced number of resonators. As the bulk of computation happens in resonators, it could reduce the time to compute. However, experiments indicated significant loss of accuracy. Possible reason for this is that filtering the pixels reduces the information content and hence it takes longer for ARN to learn. This results in reduced accuracy. To further illustrate the effect, consider two numbers 3 and 8. Presence of low value pixels on the left side is as critical to recognition of 3 as high valued pixels on the right side. Absence of the resonators identifying low value on left side of image will make it harder for ARN to distinguish 3 from 8.

### 3.4 Effect of Tuning on Learning

Learning in ARN varies depending on the value of $\rho$, T and training size. As it is mentioned earlier, increase in value of $\rho$, reduces the coverage and vice-versa. However, for $\rho < 2$ the resonance curve shown in Figure 6 is almost flat. It means that there would be very few nodes created and coverage of one node may overlap with other, leading to multiple recognitions. This scenario reflects early stages of learning when the image is identified but not properly classified. As more learning occurs, resonance gets sharper and more images will get identified. For $\rho > 3$, the resonance curve gets very narrow. Lower coverage means more failures during test. Therefore, the network starts adding large number of nodes with small coverage to attain a reasonable overall result. Recognition accuracy for these values of $\rho$ will be smaller (~85%). Threshold also has a similar effect on learning. When the threshold is low, the coverage is large and hence most of the images are recognized but because of overlap of coverage, they will not be correctly identified. Increasing the threshold will reduce the coverage and increase the number of nodes in the ARN layers. Half power point is a reasonable threshold to start with (sections 2.6 & 2.7). For improved performance, the threshold has to be increased slowly. Very high threshold reduces the coverage and thereby increasing the number of nodes. Threshold of around 0.9 (when peak is normalized to 1) yields good results.

As indicated in earlier sections, choice of resonance control parameter is data dependent and therefore needs experimentation with any given dataset. However, the results presented here give reasonable values for good learning (see Figure 7 and Figure 8). Learning within a node happens by improving the coverage. Both dilation and contraction of coverage have their use. Dilation of coverage generalizes recognition while contraction creates a more specialized recognition.

### 3.5 Ambiguity Resolution

There are three possible cases of recognition viz., Correct recognition, Wrong recognition and Multiple recognition. The case of multiple recognition needs some attention. At the output of every layer, only one winner is expected. However, there will be cases when two or more nodes produce same output at times because of learning and numerical inaccuracies. There are two possible approaches to resolve such ambiguities.

If the incoming input shifts the statistical mean towards the new input, resonance of such node may be shifted towards new input either by supervisory action of trainer or by reinforcement. This will reduce the distance between the peak and input, resulting in higher output from the node. Alternately, if the standard deviation of input is increasing, one of the nodes can relax the resonance control parameter to produce a slightly higher output. Relaxing the resonance will increase the coverage.

Resolving the ambiguity in recognition is best implemented as a wrapper around every layer. The wrapper provides local feedback and ensures that only one winner is presented at all times. Functionality of such wrapper is best implemented as a small function that suggests local corrective action. In the current implementation, it is completely supervised.

### 3.6 Explainable Paths in ARN

In a multi layer ARN, each layer represents a level of abstraction. Each layer is independent and implements a partial recognition function. There is only one winner at every layer for any input (see section 3.5). Output from a layer may be accumulated over several iterations over input space, reshaped and applied as input to next layer. Therefore, a relation between the sequentially firing neurons is input to the nodes in higher layer. This relation can be causal, temporal, spatial or a combination, depending on the function of the layer. The interface between two layers essentially records delays in the output from a lower layer and presents it as input to next layer(s). As the input to higher layers is a firing sequence, learning in ARN is similar to Hebbian learning that occurs in biological systems.

Further, output of a multi layer ARN can be traced across the layers up to the primary input layer, through a specific set of nodes and a temporal sequence. A node in every layer identifies a specific part of the whole recognition. An interface detects a relation. Therefore, every recognition event is caused by a specific path from primary input up to the output layer, with as many segments as the number of layers in ARN. This segmented path is unique to every recognition event. Therefore, input to the output can be easily interpreted in terms of partial recognitions at every layer. The path also provides a justification for the output.

### 4. RESULTS AND DISCUSSIONS

A network may be considered as robust if following conditions are met: (a) Output is range bound and does not drift, oscillate, overshoot or vanish and (b) Network recognizes the input in the presence of noise. ARN is inherently robust because (a) output of a node or layer is always range bound and (b) resonance characteristics gives noise tolerance. It is possible to add any number of feed forward layers without affecting the network stability. Further, the path taken for any specific recognition is always uniquely identifiable. This leads to a better understanding of the network and helps in performance improvement.

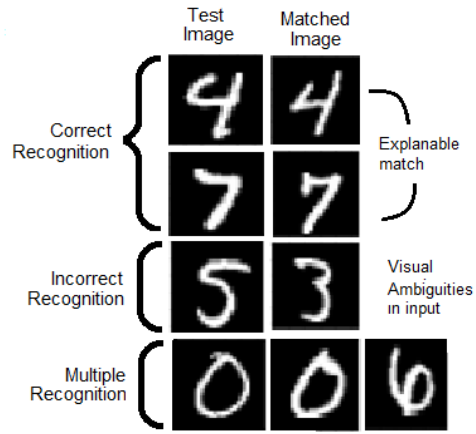

**Figure 10.** Capability of ARN to identify visually similar digits

Three possible cases of recognition are shown in Figure 10. The first two rows show the images with correct recognition. Though the matched images are fairly different from the input, ARN matches the image correctly, demonstrating the ability of ARN to select visually similar images. Images in third and fourth row are different digits but the large part of the images is very similar, emphasizing the ability of ARN to perform visual comparison. Methods to resolve ambiguities in recognition are discussed in section 3.5.

As it is discussed in earlier sections, the performance of ARN depends on $\rho$, T and sample size. It is noticed that, for a sample size of 50 per digit, the recognition is about 93% at $\rho = 2$, defined by half power point. Also at T = 0.85, accuracy of 91% is noted. As we increase the sample size, the recognition rate also increases. The learning curves of two-layer ARN for MNIST image recognition are shown in Figure 7 and Figure 8.

The confusion matrix for training sample size =200x10, $\rho$=2.42 and T=0.9 is given in Figure 11 as an illustration. Notice that digit 1 has highest true positive and 3 has lowest true positive, indicating that a higher layer is required to resolve issues with recognition of digit 3.

|   | 0 | 1 | 2 | 3 | 4 | 5 | 6 | 7 | 8 | 9 | Total |
|---|---|---|---|---|---|---|---|---|---|---|---|
| 0 | 41.75 | 1.00 | 2.50 | 2.75 | 0.00 | 3.75 | 2.00 | 0.00 | 6.25 | 0.00 | 60 |
| 1 | 0.00 | 59.00 | 0.00 | 0.00 | 0.00 | 0.00 | 0.00 | 1.00 | 0.00 | 0.00 | 60 |
| 2 | 1.31 | 4.14 | 35.97 | 8.03 | 1.61 | 2.11 | 1.11 | 2.61 | 1.98 | 1.11 | 60 |
| 3 | 0.00 | 6.17 | 3.33 | 25.83 | 0.333 | 13.33 | 1.17 | 5.00 | 4.50 | 0.33 | 60 |
| 4 | 2.00 | 5.92 | 1.42 | 0.00 | 37.58 | 0.17 | 0.16 | 2.17 | 0.48 | 10.16 | 60 |
| 5 | 3.25 | 1.70 | 0.70 | 3.25 | 1.67 | 37.95 | 3.50 | 0.20 | 4.78 | 3.00 | 60 |
| 6 | 3.19 | 1.36 | 2.20 | 0.00 | 4.69 | 2.36 | 44.30 | 0.11 | 1.61 | 0.11 | 60 |
| 7 | 0.00 | 4.03 | 0.00 | 1.33 | 3.20 | 0.00 | 0.00 | 40.03 | 0.70 | 10.70 | 60 |
| 8 | 5.37 | 2.50 | 5.20 | 2.70 | 1.58 | 4.37 | 1.20 | 1.58 | 32.41 | 3.08 | 60 |
| 9 | 0.25 | 1.00 | 0.00 | 1.25 | 8.67 | 0.25 | 0.00 | 2.83 | 2.58 | 43.16 | 60 |

**Figure 11.** Confusion Matrix for $\rho$=2.42, T=0.9, training sample size=200x10 & test sample size=60x10

### 5. COMPARATIVE PERFORMANCE OF DIFFERENT NETWORKS ON MNIST IMAGE RECOGNITION

One of the main advantages of ARN based models is the small size of training dataset to achieve reasonable accuracies. In comparison, most of the existing DNN models require large dataset to achieve similar performance. Smaller dataset also reduces the size of network, which in turn reduces the computational time and resource requirements.

A comparison of several DNNs for image recognition with MNIST dataset is available in literature [17]. The paper lists accuracies of various DNNs for 25, 50, 75 and 100 percent of MNIST dataset of 60,000 images, corresponding to 15, 30, 45 and 60 thousand training images. As expected, the recognition accuracy increases as the size of training dataset increases. At 25% of full dataset, the reported accuracies for all DNNs reported in [17] are below 88%. All the tests reported in [17] used GPU support.

As mentioned in earlier sections, we have used maximum of 5000 samples (at 500 samples per digit), i.e., only 9% of MNIST images to achieve accuracies above 93%. All our computations were done on an entry level Pentium Xeon server with 4GB RAM and no GPU support. Therefore, the performance of the new structure exceeds that of others in terms of accuracy of recognition as well as the number of samples required to achieve that accuracy.

## 6. Conclusion and Future Scope

ARN looks like a promising solution to some problems in modern Artificial Intelligence (AI). It has been used successfully in robotic path planning. This work shows its ability to perform visual comparison. The network is also being used for other biomedical applications that need image processing.

It is also possible to implement temporal relations into ARN. One such method is described in [15]. These networks can be useful for time series prediction and natural language processing. These applications are currently under development.

Fast parallel implementations of ARN with partitioned input space are possible. The problem of how features are shared among multiple partitions needs to be explored. It appears that intelligent selection of input dataset can yield better performance. Effect of adding feature neighborhoods in a multi-dimensional output of ARN needs to be explored.

Essentially, ARN is an approximating network. Therefore use of low precision arithmetic for ARN is natural to its implementations. A 16 bit fixed point representation is being explored [22] for use in ARN. Use of low precision arithmetic in other structures like spiking neural networks [23] has also been reported in literature. Coverage of ARN node intrinsically supports quantization of input space and hence use of lower precision in number representation has little effect on accuracy but greatly reduces the computational time.

## Acknowledgment

Part of this work was reported by the authors in a recent conference [24]. The authors would like to thank C-Quad Research, Desur IT Park, Belagavi for all the facilities and support provided. We would also thank our respective institutions viz., KLE Dr M S Sheshgiri Collge of Engineering & Technology, Belagavi, Siddaganga Institute of Technology, Tumakuru, and Visvesvaraya Technological University, Belagavi.

## AUTHOR BIOGRAPHY

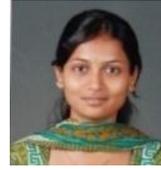

**Shilpa Mayannavar** is a Research Scholar at C-Quad Research, Belagavi, Karnataka, India. She has obtained Bachelor of Engineering in Electronics and Communication Engg. (2012) and Master of Technology in VLSI Design and Embedded System (2014) from Visvesvaraya Technological University (VTU), Belagavi. She has one year of teaching experience and three years of research experience. Her research interests are Processor design, Artificial Intelligence and Neural Networks.

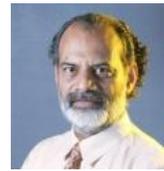

**Uday Wali** is a Professor in Dept. of ECE at KLE Dr M S Sheshgiri College of Engineering. & Technology, Belagavi, Karnataka India. He has obtained Bachelor of Engineering in Electrical and Electronics Engg. from Karnataka University Dharwad (1981) and Ph.D from IIT Kharagpur (1986). He is a fellow of Institute of Engineers (India) and CEO of C-Quad Computers, Desur IT Park, Belagavi. He has 30 years of teaching and 15 years of industrial experience. His current research interests are Artificial Intelligence, Neural Networks, and Processor Design.

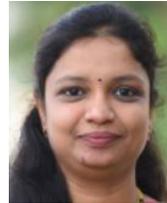

**V M Aparanji** is an Associate Professor at Siddaganga Institute of Technology, Tumakuru. She has completed Bachelor of Engineering (2003) and Master of Science by Research (2010). She has obtained Doctor of Philosophy in Neural Networks and Artificial Intelligence from VTU, Belagavi (2018). She has 15 years of teaching experience. Her areas of Interests are Embedded systems, Neural Networks, Robotics, Artificial Intelligence and Machine Learning.